\documentclass[sigconf]{acmart}
\usepackage{multirow}
\usepackage{multicol}
\usepackage{latexsym}
\usepackage{amsmath}
\usepackage{graphicx}
\usepackage[caption = false]{subfig}
\usepackage{booktabs}

\AtBeginDocument{%
  \providecommand\BibTeX{{%
    \normalfont B\kern-0.5em{\scshape i\kern-0.25em b}\kern-0.8em\TeX}}}

\copyrightyear{2021}
\acmYear{2021} 
\setcopyright{acmcopyright}
\acmConference[CIKM '21]{Proceedings of the 30th ACM International Conference on Information and Knowledge Management}{November 1--5, 2021}{Virtual Event, QLD, Australia}
\acmBooktitle{Proceedings of the 30th ACM International Conference on Information and Knowledge Management (CIKM '21), November 1--5, 2021, Virtual Event, QLD, Australia}
\acmPrice{15.00}
\acmDOI{10.1145/3459637.3481924}
\acmISBN{978-1-4503-8446-9/21/11}

\begin{document}
\fancyhead{}

\title{GEDIT: Geographic-Enhanced and Dependency-Guided Tagging for Joint POI and Accessibility Extraction at Baidu Maps}


\author{\texorpdfstring{Yibo Sun$^{1}$}{Yibo Sun}, 
        \texorpdfstring{Jizhou Huang$^{1\dagger}$}{Jizhou Huang}, 
        \texorpdfstring{Chunyuan Yuan$^{1}$}{Chunyuan Yuan},
        \texorpdfstring{Miao Fan$^{1}$}{Miao Fan},
        \texorpdfstring{Haifeng Wang$^{1}$}{Haifeng Wang},
        \texorpdfstring{Ming Liu$^{2}$}{Ming Liu},
       \texorpdfstring{Bing Qin$^{2}$}{Bing Qin}
       }
\thanks{$^\dagger$Corresponding author: Jizhou Huang.}
\affiliation{%
  \institution{$^{1}$Baidu Inc., Beijing, China}
  \institution{$^{2}$Research Center for Social Computing and Information Retrieval, Harbin Institute of Technology, China}
}
\email{{sunyibo, huangjizhou01, yuanchunyuan, fanmiao,wanghaifeng}@baidu.com,{mliu, qinb}@ir.hit.edu.cn}

\renewcommand{\shortauthors}{Sun et al.}

\begin{abstract}
Providing timely accessibility reminders (such as closed and relocated) of a point-of-interest (POI) plays a vital role in improving user satisfaction of finding places and making visiting decisions.
However, it is difficult to keep the POI database in sync with the real-world counterparts due to the dynamic nature of business changes and innovations.
To alleviate this problem, we formulate and present a practical solution that jointly extracts POI mentions and identifies their coupled accessibility labels from unstructured text (hereafter referred to as \textit{joint POI and accessibility extraction}). 
We approach this task as a sequence tagging problem, where the goal is to produce <POI name, accessibility label> pairs from unstructured text.
This task is challenging because of two main issues: (1) POI names are often newly-coined words so as to successfully register new entities or brands and (2) there may exist multiple pairs in the text, which necessitates dealing with one-to-many or many-to-one mapping to make each POI coupled with its matching accessibility label.
To this end, we propose a \textbf{G}eographic-\textbf{E}nhanced and \textbf{D}ependency-gu\textbf{I}ded sequence \textbf{T}agging (\textbf{GEDIT}) model to concurrently address the two challenges.
First, to alleviate challenge \#1, we develop a geographic-enhanced pre-trained model to learn the text representations, which is able to significantly relieve the problem of newly-coined words.
Second, to mitigate challenge \#2, we apply a relational graph convolutional network to learn the tree node representations from the parsed dependency tree, which enables us to establish a correlation between a POI and its accessibility label.
Finally, we construct a neural sequence tagging model by integrating and feeding the previously pre-learned representations into a CRF layer.
Extensive experiments conducted on a real-world dataset demonstrate the superiority and effectiveness of GEDIT.
In addition, it has already been deployed in production at Baidu Maps, and it successfully keeps processing hundreds of thousands of Web documents every week. 
Statistics show that the proposed solution can save significant human effort and labor costs to deal with the same amount of documents, which confirms that it is a practical way for POI accessibility maintenance.
\end{abstract}


%


\maketitle

\section{Introduction}
\label{sec:intro}
\begin{figure}[!t]
	\centering
	\includegraphics[width=1.0\linewidth,trim={0.2cm 0.85cm 0.2cm 0.1cm},clip]{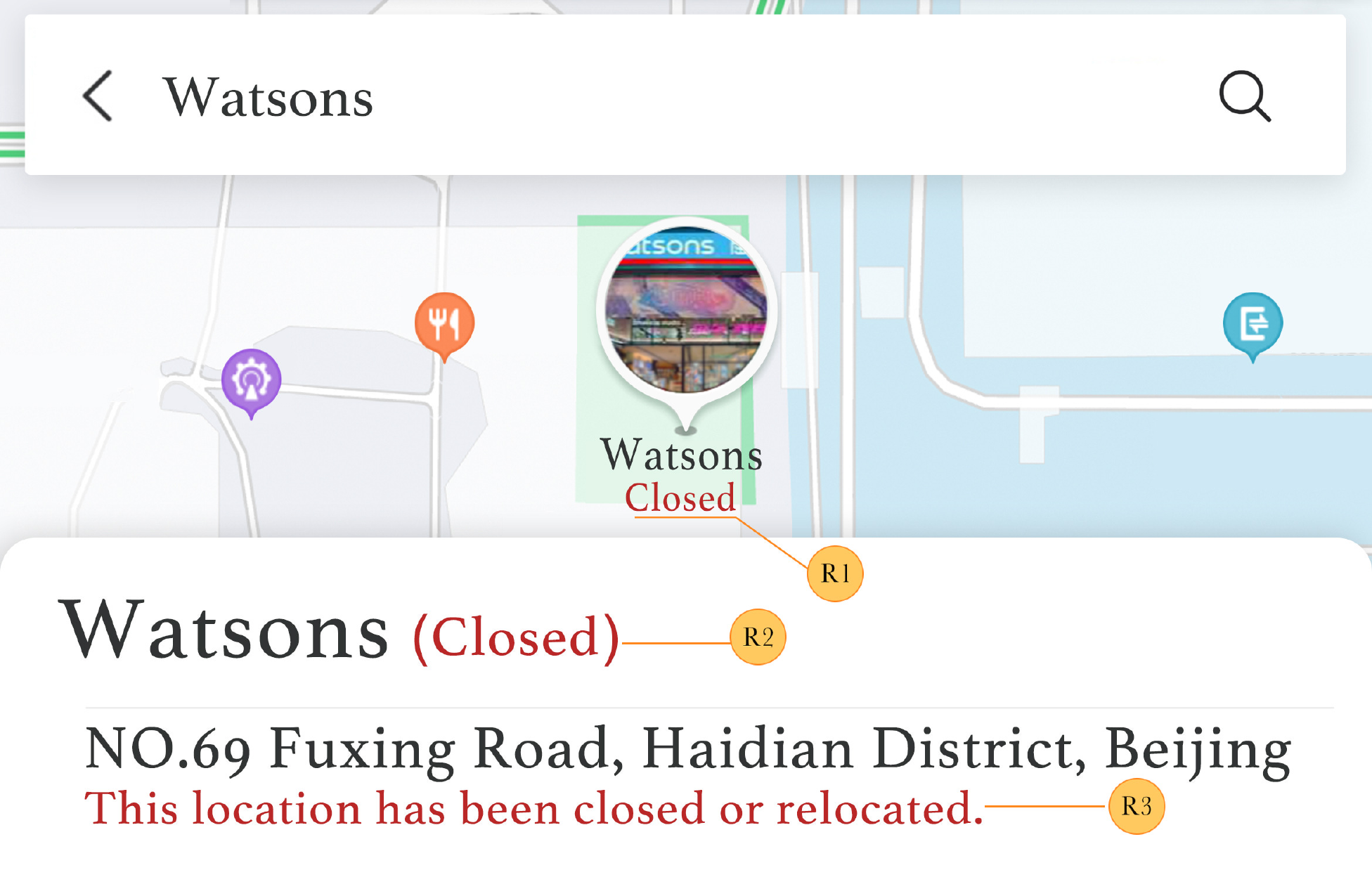} 
	\caption{Example of POI accessibility reminders (i.e., R1, R2, and R3) at Baidu Maps.}
	\label{fig:product-case}
\end{figure}
In commercial map applications such as Baidu Maps, rich and timely updated POI information (such as POI address, POI coordinates, and POI accessibility reminder) plays an important role in enabling users to entertain location-based services.
Among which, the accessibility reminder is of vital importance to users since it is frequently used to support decisions when planning to visit a POI.
Figure~\ref{fig:product-case} shows an example of the POI multidimensional information page at Baidu Maps. It can be seen that we have strongly prompted the closing status of the POI at three different places. We hope that users can be fully informed that the POI has been closed before they decide to visit it, so as to avoid the disappointments experienced after traveling tens of kilometers farther to it.
Therefore, in order to make sure that the users suffer as little inconvenience as possible when finding places or making visiting decisions with a map application, it is important to provide timely accessibility reminders.

However, it is difficult to keep the POI database in sync with the real-world counterparts due to the dynamic nature of business changes and innovations. 
Statistics show that $74.5\%$ of the POIs at Baidu Maps have been updated in 2020. 
It is extremely time-consuming and expensive to handle such a large number of updates if we heavily rely on human efforts. 
To reduce labor costs and increase productivity, several recent work has attempted to develop new ways to maintain a POI database.
\citet{Revaud2019DidIC} proposed to use street-view images to automatically detect changes of POIs. 
Although it is feasible, the acquisition of geo-tagged street-view images at different times is time-consuming and expensive, which limits its practical applicability when applying to update large-scale POIs.
In addition, several researchers proposed to extract POI names from text~\cite{Rae2012MiningTW,Chuang2018DetectingOP,Xu2019DLocRLAD}.
Although extracting POI names is the first and important step towards the maintaining of POI information, there are indispensable attributes that need to be extracted and correlated with the corresponding POIs.
Nevertheless, new POIs emerge endlessly and their names are often newly-coined words, while the existing POIs are subject to change over time, resulting in a higher uncertainty on the task frequency and cost. Therefore, it is critical to explore more effective ways to jointly detect POIs and extract their associated attributes from text.

\begin{figure}[!th]
	\centering
	\includegraphics[width=\linewidth,trim={0cm 0.9cm 0cm 1.0cm},clip]{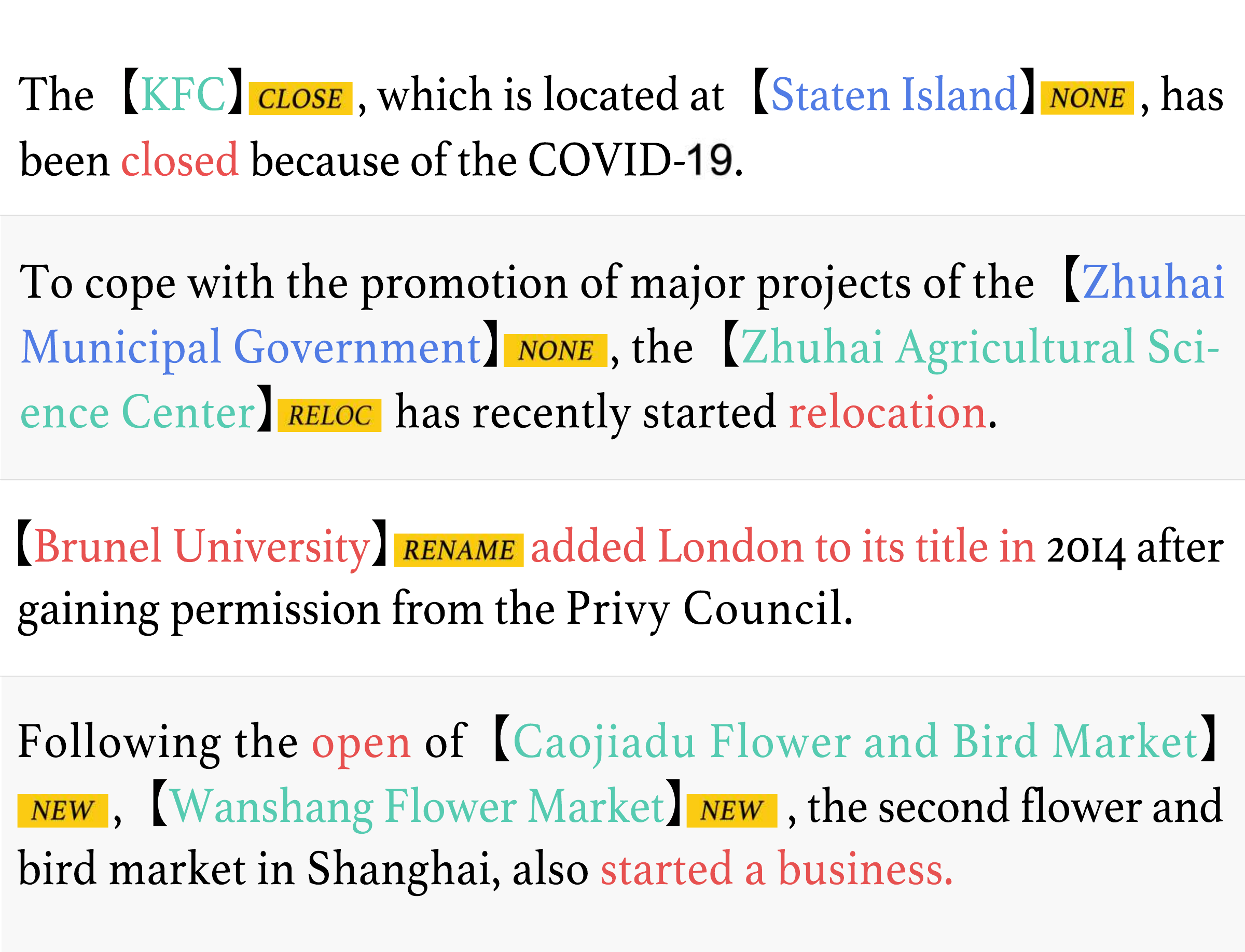}
	\caption{Examples of joint POI and accessibility extraction from unstructured text.}
	\label{fig:task-definition}
\end{figure}

After meticulous analysis, we find that many business entities prefer to publish the business change information on their official websites or Internet news in a timely fashion. This demonstrates that massive Web pages are valuable data sources for large-scale extraction of POI change information. As POI accessibility is vital information to users, we present a practical solution that jointly extracts POI mentions and identifies their coupled accessibility labels from unstructured text (hereafter referred to as \textit{joint POI and accessibility extraction}).
We frame this task as a sequence tagging problem and consider the following four mainstream accessibility labels: one for emerging POIs: \textbf{NEW} and the other three for updating the accessibility of existing POIs: \textbf{RENAME}, \textbf{RELOC} (the abbreviation of ``relocation''), and \textbf{CLOSE}. 
Figure~\ref{fig:task-definition} shows four representative examples of this task.
This task is challenging because of the following two main issues.

(1) \textbf{Rare or unknown words:} POI names are often newly-coined words so as to successfully register new entities or brands. As a result, POI names are typically regarded as out-of-vocabulary (OOV) words, the semantic meaning of them can hardly be captured by neural-based models. 
As illustrated by the first example in  Figure~\ref{fig:task-definition}, KFC is a well-known chain brand, which widely exists in our POI database. However, \textit{Staten Island} is absent from our POI database, which will be regarded as an OOV word.  

(2) \textbf{Many-to-one or one-to-many mapping:} There may exist multiple <POI name, accessibility label> pairs in the text, which necessitates dealing with one-to-many or many-to-one mapping to make each POI coupled with its matching accessibility label.
For example, the first sentence in Figure~\ref{fig:task-definition} mentions two POI names (KFC and Staten Island), but there exists only one accessibility label (CLOSE). 
Therefore, the following two pairs should be extracted from it, i.e., <KFC, CLOSE> and <Staten Island, NONE>.

To this end, we propose a \underline{\textbf{G}}eographic-\underline{\textbf{E}}nhanced and \underline{\textbf{D}}ependency-gu\underline{\textbf{I}}ded \underline{\textbf{T}}agger (\textbf{GEDIT}) to concurrently address the two challenges. GEDIT casts the POI accessibility recognition task as a sequence tagging problem by giving each token a joint  mention-accessibility label. As a result, GEDIT is able to jointly extract POI mentions and identify their accessibility labels. 
Consequently, GEDIT can produce arbitrary number of <POI name, accessibility label> pairs simultaneously. 

To alleviate challenge \#1, GEDIT adopts a geographic-enhanced pre-trained language model~\cite{Devlin2019BERTPO}, which is able to significantly relieve the problem of newly-coined POI names. For example, by taking advantage of the geographic knowledge in the addresses of existing POIs, the pre-trained model is able to learn the patterns of coining new POI names. As a result, new POIs could be better handled.

To mitigate challenge \#2,
we apply a relational graph convolutional network (RGCN) \cite{schlichtkrull2018modeling} to learn the tree node representations from the parsed dependency tree, which enables us to establish a correlation between a POI and its accessibility label.
As a result, GEDIT is able to avoid the distraction from the auxiliary POIs that do not have any accessibility changes. 
Take the first sentence in  Figure~\ref{fig:task-definition} as an example, with the aid of rhetorical relation between the word ``closed'' and ``KFC'', it is easy to know that the closed POI is ``KFC'' rather than ``Staten Island''.

Finally, we construct a neural sequence tagging model by integrating and feeding the previously pre-learned representations into a CRF \cite{Lafferty2001ConditionalRF} layer.

\begin{figure*}[htbp]
	\centering
	\includegraphics[width=1.0\textwidth]{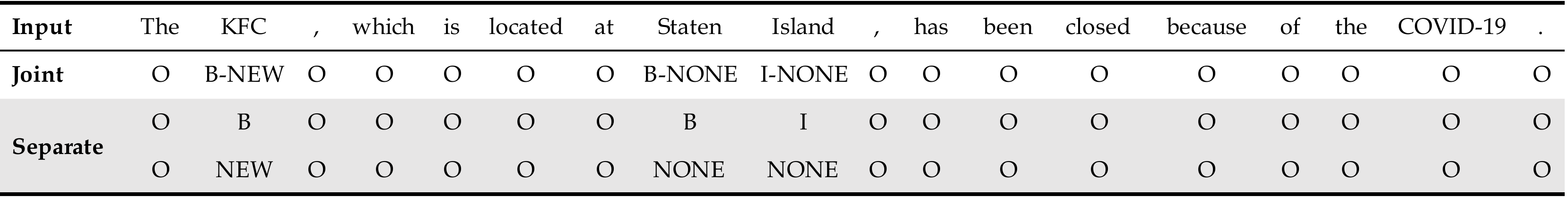}
	\caption{Labeling schema.}
	\label{fig:label-schema}
\end{figure*}

Given the lack of an appropriate benchmark, we construct and release a large-scale real-world dataset named WebPOIs.\footnote{The dataset is publicly available at~\url{https://github.com/PaddlePaddle/Research/tree/master/ST_DM/CIKM2021-GEDIT/}}
Extensive experiments conducted on WebPOIs dataset demonstrate that GEDIT significantly outperforms several strong sequence tagging baselines with a large margin.
Statistics show that the proposed solution can save significant human effort and labor costs to deal with the same amount of documents, which confirms that it is a practical way for POI accessibility maintenance.

Our contributions can be summarized as follows:
\begin{itemize}
    \item \textbf{Potential impact}: Geographic-enhanced and dependency-guided sequence tagging (GEDIT) model is our first attempt to devise a neural model that handles large-scale text  to maintain the accessibilities of hundreds of millions of POIs at Baidu Maps. GEDIT has been successfully deployed in production at Baidu Maps.  It keeps inspecting hundreds of thousands of documents every week, saving significant labor costs in practice.
    
    \item \textbf{Novelty}: The design and implementation of GEDIT are driven by the novel idea that takes advantage of a geographic-enhanced pre-trained model and dependency relations to guide the sequence tagging model, which is able to produce more accurate results from text.

    \item \textbf{Technical quality}: The offline experiments demonstrate that GEDIT can consistently achieve significant improvements on $F_1$ score in comparison with several strong baselines. After we deployed GEDIT in production, the efficiency of manual verification increases by 17.8\%, which dramatically saves the maintenance costs at Baidu Maps.
    \item \textbf{A new and challenging dataset}:
    The WebPOIs dataset is composed of 19,333 documents and 99,139 POIs, which is expected to bring this substantial but challenging task to the attention of researchers both in academia and in industry.
\end{itemize}

\section{Task Formulation and Dataset}
\subsection{Task Formulation}
\label{sec:task_formulation}
Given a document of $n$ words, denoted by $D = \{x_1,\dots,x_n\}$, the outputs of the proposed task are all continuous sub-sequence chunks $p_i = \{x_i,\dots,x_j\}$ representing POIs and their accessibility label $s_i \in \{$NEW, RELOC, RENAME, CLOSE$\}$.

We first break the task into two sub-tasks, including \textbf{POI term extraction} (PTE) for extracting POIs from the document and \textbf{POI accessibility identification} (PAI) for identifying the accessibility of each POI term. Instead of performing PTE and PAI step by step in a pipeline paradigm that does not fully exploit the joint information between them, our proposed framework learns the two sub-tasks jointly. Since PTE is a sequence tagging task and PAI is a classification task, they cannot be directly trained together. Thus, we convert PAI to a sequence tagging task by giving each POI token an accessibility label.

With the help of the sequence decision in PAI that models the relationship of each POI accessibility, this joint training paradigm could learn to extract arbitrary pairs of <POI name, accessibility label> from the text efficiently.
We use BIO schema~\cite{Ratinov2009DesignCA} where the prefixes B, I, and O indicate the \textbf{B}eginning, the \textbf{I}nside, and the \textbf{O}utside of a chunk, respectively.
To exploit the effectiveness of different labeling schemas, we design two different labeling settings in this work, i.e., joint setting and separate setting. 
As shown in Figure~\ref{fig:label-schema}, the joint setting represents the information for POI and its accessibility simultaneously in one label set. By contrast, the separate setting uses two kinds of labels as two sub-tasks. Formally, for each word $x_i$, in the separate setting, we assign a tag $t^p_i \in T^p$ in PTE and assign a tag $t^s_i \in T^s$ in PAI, where $T^p = \{$B-POI, I-POI, O$\}$ and $T^s = \{$NEW, RELOC, RENAME, CLOSE, NONE, O$\}$. In the joint setting, we integrate $T^p$ and $T^s$ into one set $\{$B-NEW, I-NEW, B-RELOC, I-RELOC, B-RENAME, I-RENAME, B-CLOSE, I-CLOSE, B-NONE, I-NONE, O$\}$.

\subsection{Benchmark Dataset}
\label{sec:data_collection}
As there is no public dataset available for this task, we construct the WebPOIs dataset using a three-step way.

\paragraph{Document Collection}
To retrieve high-quality documents containing POI information, we utilize multiple data sources, including general Web documents and official websites. We manually construct queries for searching the public Web documents and keep the top-ranked results returned by a search engine. For the websites, we crawl the documents based on a list of keywords related to POI accessibility.

\paragraph{Pruning} After document collection, we use a two-step pruning operation to ensure that the obtained documents contain high-quality POI accessibility information.
First, we prune the documents based on the number of POIs recognized by a pre-trained POI recognizer.
We only keep those documents that contain two or more POIs detected by the POI recognizer.
The POI recognizer is a high-performance sequence tagging model that has been deployed in Baidu Maps.
We conduct this step to make sure that the documents convey information related to POIs.
Second, we prune the documents based on a dictionary containing words that express the meaning of POI accessibility, such as ``move'' and ``open''.
We keep those documents that contain at least one word in the dictionary.
This step is conducted to make sure that the documents contain descriptions of POI accessibility.

\paragraph{Human Annotation}
To annotate a high-quality dataset, we hire a team of full-time annotators, and select qualified annotators using the following process.
First, the annotators are told to learn a carefully crafted annotation guideline that includes examples of excellent and lousy POI accessibility change annotations on those documents and why they were categorized as such.
Then, they are told to practice and examine themselves on two small sets of documents with correct labels.
This train-practice-examine process is iterated three times within a month.

After selecting the qualified annotators, we separate them into an annotation team and a quality assurance (QA) team. We first ask the annotation team to mark all POIs and their accessibility labels including \textbf{NEW}, \textbf{RENAME}, \textbf{RELOC}, \textbf{CLOSE}, and \textbf{NONE}. 
Then, we ask the QA team, which consists of annotators with the top scores in the examination, to inspect over 20\% of the labeled data randomly. Finally, the researchers randomly inspect 5\% of the labeled data and merge the data into WebPOIs if they are clean enough. If the accuracy is below 90\%, the whole data are sent back to re-annotate in each inspection stage. 
For each annotation-inspection process, we process 1,000 documents. All the procedures mentioned above are conducted on an in-house CMS system. To motivate the annotators to perform high-quality annotations, the higher the annotation accuracy achieves, the more they are paid.

The WebPOIs dataset comprises 19,333 documents and 99,139 POIs. Table \ref{table:data-stat} and Table \ref{table:data-stat-label} show the detailed statistics of WebPOIs.
This new dataset enables us to analyze POIs that appeared within complex linguistic phenomena.

\begin{table}[!ht]
	\centering
	\caption{Statistics of WebPOIs.}
	\begin{tabular}{lr}
		\toprule
		\textbf{Item} & \textbf{Number} \\
		\midrule
		\midrule
		\# of Document & 19,333 \\
		\# of POI & 99,139 \\
		\# of unique POI & 44,167 \\
		Averaged POIs / Document & 5.1 \\
		Averaged Words / Document & 195.7 \\
		Averaged Words / POI & 7.1 \\
		\bottomrule
	\end{tabular}
	\label{table:data-stat}
\end{table}

\begin{table}[!h]
	\centering
	\caption{The number of POIs of each label in WebPOIs.}
	\begin{tabular}{lrrrr}
		\toprule
		\textbf{LABEL} & \textbf{Train} & \textbf{Dev} & \textbf{Test} & \textbf{Total} \\
		\midrule
		\midrule
		\# NONE   & 37,488 & 5,130 & 10,874 & 53,492 \\
		\# NEW    & 17,266 & 2,372 & 4,956 & 24,594 \\
		\# CLOSE  & 8,565 & 1,258 & 2,299 & 12,122 \\
		\# RELOC  & 4,637 & 679 & 1325 & 6,641 \\
		\# RENAME & 1,609 & 229 & 452 & 2,290 \\
		\bottomrule
	\end{tabular}
	\label{table:data-stat-label}
\end{table}

\section{GEDIT}
\label{sec:method}
 
\begin{figure*}[!ht]
	\centering
	\includegraphics[width=1.\linewidth,trim={0.8cm 0.2cm 0.8cm 0.7cm},clip]{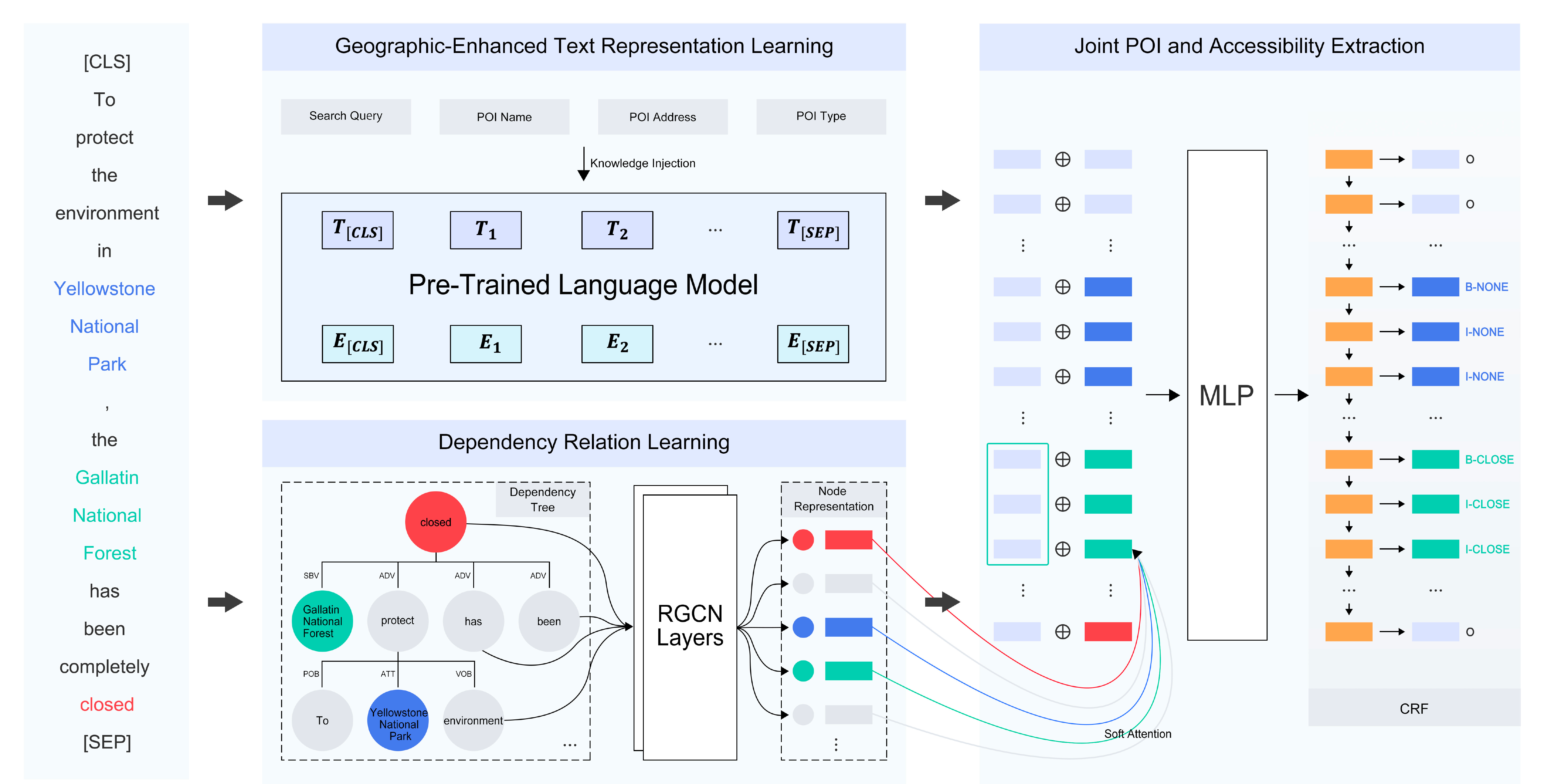}
	\caption{The overall framework of GEDIT. Given an input document $D$, we first obtain the sub-word representations with GERNIE and its dependency tree node representations with RGCN, respectively. Then, we use an attention mechanism to get $D$'s fused representations.
The fused representations are finally used to jointly extract POI and accessibility labels.}
	\label{fig:framework}
\end{figure*}

In this section, we detail the proposed model \textbf{GEDIT}. As shown in Figure~\ref{fig:framework}, GEDIT contains three major components:
(1) Geographic-Enhanced Text Representation Learning, (2) Dependency Relation Learning, and (3) Joint POI and Accessibility Extraction.
For an input document $D$, we first use component \#1 to learn the geographic-enhanced text representations of $D$.
Simultaneously, we use component \#2 to learn the dependency tree node representations of $D$.
Finally, we use component \#3  to get $D$’s fused representations, and then jointly extract POI and accessibility labels by using the fused representations.

\subsection{Geographic-Enhanced Text Representation Learning}
To relieve the problem of newly-coined POI names, we explore to incorporate prior geographic knowledge into a pre-trained language model ERNIE~\cite{Sun2019ERNIEER}.
The geographic knowledge comes from the massive POI database and the POI search logs at Baidu Maps.
Specifically, we incorporate geographic knowledge into the model by continuing to train a mask language model (MLM) task based on the parameters of ERNIE. 
In the MLM task, we organize each document in the form of the concatenation of the following four types of text information: (1) the most frequent query when searching for a POI, (2) the full POI name, (3) the POI address, and (4) the POI type.
We separate the query from other information with a [SEP] token.
We use the whole word mask (WWM) strategy to make predictions for the phrases in each document. We use a query component analysis module deployed at Baidu Maps to split each document at the granularity of geographic entities.
Each geographic entity in a document has a 15\% probability of being masked and predicted by the language model during the training process.
For each word in the selected entity, we replace the word with a ``[MASK]'' token with 70\% probability, replace the word with a misspelled word with 10\% probability, replace the word with a random word with 10\% probability, and leave the word unchanged with 10\% probability.
The words in a query that do not match any words in the target POI name are treated as misspelled words.
With this training procedure, we can learn four types of geographic knowledge in the MLM task as follows.
(1) The natural language description of POI name and address. 
(2) The relationship between POI name, address, and type. 
(3) The relationship between query, POI name, and address.
(4) The possible misspelling of POI name and address.

Formally, given a document $D$, we first tokenize $D$ into a sub-word sequence $\{s_1,\dots,s_i, \dots, s_L\}$, where $s_i$ denotes each sub-word and $L$ represents the length of the sub-word sequence. Then we use the above-mentioned geographic-enhanced ERNIE (GERNIE) to get the sequence of latent vectors $\{\mathbf{s}_1, \dots, \mathbf{s}_L\} = GERNIE(\{s_1,\dots,s_L\})$. These latent vectors are shared among PTE and PAI.

\subsection{Dependency Relation Learning}
Not every POI in a document is followed by an accessibility change. Such POIs would significantly confuse the model to identify which POIs have accessibility changes. 
Based on observations, some words can indicate the accessibility changes, which inspires us to link these kinds of words to the target POI to facilitate determining its accessibility label. 
The dependency tree of the document can reflect the rhetorical relations of different nodes, which could help link these kinds of indicator words to the target POI nodes. Thus, we consider encoding the dependency tree into the text representations.

\subsubsection{Dependency Tree Construction}
Specifically, we first segment the document into sub-words. Then, we use a dependency parsing tool\footnote{\url{https://github.com/baidu/DDParser}} to construct the dependency tree $\mathcal{G}(V, E)$ of the given document, where $V$ is the node-set of the dependency tree, and $E$ is dependency relations set. In this paper, we use 14 pre-defined rhetorical relations $\mathcal{R}$ as the type of the tree edge. Finally, we record the mapping relations $\mathcal{R}^{map}$ between the sub-words of the original document and the tree nodes.

\subsubsection{Dependency Relation Learning}
The dependency tree contains different types of relations, and various relations may play different roles in generating node representations. For example, the subject-verb relation and the verb-object relation may be more important than other kinds of relations. To thoroughly learn this difference, we apply a relational graph convolutional network (RGCN)~\cite{schlichtkrull2018modeling} to encode the dependency tree and learn the node representations.

First, we use the average of the sub-word representations as the initial node features of the RGCN input. For each node $x_i$, the initial representation of it is computed as follows:
\begin{equation}
\mathbf{x}_i = \mathop{mean}([\mathbf{s}_{i+j}, \ldots, \mathbf{s}_{i+k}])    \,,  \\
\end{equation}
where the node $x_i$ is composed of sub-words $s_{i+j}, \ldots, s_{i+k}$.

Then, we utilize the RGCN to encode the structure information into the node representations. Unlike regular GCNs, the RGCN introduces relation-specific transformations, i.e., depending on the type and direction of an edge, and accumulating transformed feature vectors of neighboring nodes through the type of the edge. This process is formulated as follows:
\begin{equation}
	h_{i}^{(l+1)} = \sigma \left( \sum_{r \in \mathcal{R}} \sum_{j \in \mathcal{N}_i^r} \frac{1}{c_{i,r}} \mathbf{W}_r h_{j}^{(l)} + \mathbf{W}_0 h_{i}^{(l)}  \right)  \,,  \\
\end{equation}
where $\mathbf{W}_r, \mathbf{W}_0$ are trainable matrices. $l$ denotes the $l$-th RGCN layer. $\mathcal{N}_i^r$ denotes the set of neighbor indices of node $i$ under relation $r \in \mathcal{R}$. $c_{i,r} = |\mathcal{N}_i^r|$ is a problem-specific normalization constant. We use two layers of RGCN to learn the tree structure and obtain the node representations $\mathbf{H}^{(2)} \in \mathbb{R}^{|V| \times d}$ of the dependency tree.

\subsection{Joint POI and Accessibility Extraction}
After obtaining the document $D$'s sub-word representations with GERNIE and its dependency tree node representations with RGCN, we use an attention mechanism to get $D$'s fused representations, which are used to jointly extract POI and accessibility label.

\subsubsection{Representation Fusion}
To establish an efficient connection between the words in each document and their corresponding nodes in the dependency tree, we explore two kinds of fusion strategies to produce the sub-word representations as follows.

\paragraph{Hard Attention Fusion}
For each sub-word in a given document, they all have a corresponding node in the dependency tree, recorded in mapping table $\mathcal{R}^{map}$. We fetch the node representation and concatenate it with the sub-word representation. For example, given sub-word id $s_i$, we look up the node id $n_i$ from $\mathcal{R}^{map}$, find the node representation from $\mathbf{H}^{(2)}$, and then concatenate it with sub-word representation $\mathbf{s}_i$, which is formulated as follows:
\begin{equation}
	\begin{split}
	&  n_i = \mathcal{R}^{map}[s_i]   \,, \\
	&  \widetilde{\mathbf{s}}_i = \mathbf{concat}(\mathbf{s}_i, \mathbf{H}^{(2)}_{n_i})  \,, \\
	\end{split}
\end{equation}
where $\widetilde{\mathbf{s}}_i \in \mathbb{R}^{d}$ is the final sub-word representation.

\paragraph{Soft Attention Fusion}
The hard attention fusion strategy only uses the node representation belonging to the sub-word, which cannot fully utilize the node representations with similar semantic information. Thus, we design a soft attention module to fuse all node representations as final sub-word representations. 

Specifically, given a sub-word id $s_i$, we look up its sub-word representation $\mathbf{s}_i$ from the output of GERNIE, and obtain all node representations $\mathbf{H}^{(2)} \in \mathbb{R}^{|V| \times d}$. The soft attention is defined as:
\begin{equation}
\begin{split}
& \mathbf{z} = \tanh (\mathbf{H}^{(2)} \mathbf{W} \mathbf{s}_i)  \,,  \\
& \mathbf{\alpha}_i = \frac{exp(\mathbf{z}_i)}{\sum_{j=1}^{|V|} exp(\mathbf{z}_j )}  \,,  \\
& \mathbf{s}_i^{\prime} = \sum_{k=1}^{|V|}{\mathbf{\alpha}_k \mathbf{H}^{(2)}_k}  \,, \\ 
\end{split}
\end{equation}
where $\mathbf{W} \in \mathbb{R}^{d \times d} $ is the trainable matrix and $\mathbf{\alpha} \in \mathbb{R}^{ |V| \times 1 } $ is the attention weight. $\mathbf{s}_i^{\prime}$ is the aggregated node representation of sub-word $s_i$. 

Then, we concatenate $\mathbf{s}_i$ and $\mathbf{s}_i^{\prime}$ to obtain the final sub-word representation:
\begin{equation}
\widetilde{\mathbf{s}}_i = \mathbf{concat}(\mathbf{s}_i, \mathbf{s}_i^{\prime})  . \\
\end{equation}

After obtaining the final sub-word representation by hard attention or soft attention module, we feed it into a fully connected layer to produce the final representation:
\begin{equation}
	\widehat{\mathbf{s}} = \mathbf{W}_2 \mathbf{ReLU} (\mathbf{W}_1\widetilde{\mathbf{s}}_i + b)  \,, \\
\end{equation}
where $\mathbf{W}_1$ and $\mathbf{W}_2$ are trainable matrices.  $\mathbf{ReLU}(\cdot)$ is a nonlinear activation function~\citep{agarap2018deep}.

\subsubsection{POI and Accessibility Tagging}
After obtaining the fused sub-word representations, we explore two settings: (1) separate setting and (2) joint setting, to predict the chunks of the POIs and their accessibility labels.

\paragraph{Separate Setting}
For the separate setting, we treat the chunks of the POIs and the accessibility prediction as two independent sub-tasks. Specifically, the sub-word representations $\widehat{\mathbf{S}} = [\widehat{\mathbf{s}}_1, \widehat{\mathbf{s}}_2, \ldots, \widehat{\mathbf{s}}_L ]$ is fed into two different CRF models to generate the labels for PTE and PAI, respectively:
\begin{equation}
p(y^p|D; \mathbf{\theta}^p) = \mathbf{CRF}_1(\widehat{\mathbf{S}} )  \,, \quad p(y^s|D; \mathbf{\theta}^s) = \mathbf{CRF}_2(\widehat{\mathbf{S}} )  \,,
\end{equation}
where $y^p$ is the POI sequence labels and $y^s$ is the accessibility labels. $\mathbf{\theta}^p$ and $\mathbf{\theta}^s$ denote all parameters used for both sub-tasks.

After tagging, the remaining step is to obtain the POI terms and their accessibility labels.
It is convenient to get the POI terms of the given sentence according to the meaning of the elements in $T^p$.
To generate the accessibility of each POI term, we regard the POI term as the boundary of the accessibility labels and then count the number of  accessibility labels within the boundary. We adopt a voting mechanism, regard the accessibility label with the maximum number as the ultimate label for the POI term. If there exist equal amount of accessibility labels, we regard the first one as the final result.
For example, the final label of ``RELOC, RELOC, RENAME, RENAME'' is ``RELOC'', the final label of ``NEW, NEW'' is ``NEW'', and the final label of ``NEW, RELOC, RELOC'' is ``RELOC''. This method is simple but effective.

\begin{table*}[!htbp]
	\label{table-results-compare}
	\caption{Comparison of all models in terms of $F_1$ metrics. GEDIT$_{soft}$/GEDIT$_{hard}$ denotes the GEDIT model with soft/hard attention. Improvements of GEDIT over all other models are statistically significant using $t$-test for $p < 0.01$.}
	\setlength{\tabcolsep}{2mm}{
		\begin{tabular}{l|lrrrrrr}
			\toprule
			\textbf{Settings} &  \textbf{Model}      &   \textbf{NEW $F_1$} & \textbf{CLOSE $F_1$} & \textbf{RELOC $F_1$} & \textbf{RENAME $F_1$} & \textbf{Macro $F_1$} & \textbf{Micro $F_1$} \\
			
			\midrule
			\midrule
			\multirow{7}{*}{\textbf{Separate Setting}}                 
			& CNN+CRF     & 0.242  & 0.320    & 0.407    & 0.395   & 0.302    & 0.295    \\
			& LR-CNN      & 0.328  & 0.407    & 0.314    & 0.275   & 0.343    & 0.344    \\
			& BiLSTM+CRF  & 0.482  & 0.480    & 0.542    & 0.511   & 0.492    & 0.492    \\
			& CNN+BiLSTM+CRF    & 0.548    & 0.562    & 0.604   & 0.631    & 0.565    & 0.565    \\
            & ERNIE+CRF & 0.597   & 0.640   & 0.688 & 0.572  & 0.620    & 0.621   \\
            \cline{2-8}
			& GEDIT$_{soft}$  & 0.633   & 0.644  & 0.687 & 0.703  & 0.648    & 0.648  \\
			& GEDIT$_{hard}$ & 0.637   & 0.647  & 0.718 & 0.693  & 0.654    & 0.654  \\
			\midrule
			\multirow{10}{*}{\textbf{Joint Setting}}                 
			& CNN+CRF     & 0.543  & 0.571    & 0.603    & 0.496   & 0.557    & 0.557    \\
			& LR-CNN      & 0.614  & 0.651    & 0.677    & 0.630   & 0.634    & 0.634    \\
			& BiLSTM+CRF  & 0.629  & 0.680    & 0.709    & 0.714   & 0.658    & 0.658    \\
			& CNN+BiLSTM+CRF    & 0.662   & 0.702   & 0.734  & 0.715  & 0.686    & 0.686    \\
			& ERNIE+CRF & 0.663   & 0.761  & 0.786 & 0.756  & 0.711    & 0.712    \\
			\cline{2-8}
			& GEDIT$_{soft}$     & 0.733   & 0.772  & 0.781 & 0.762  & 0.752    & 0.752   \\
			& GEDIT$_{hard}$     & \textbf{0.741}   & \textbf{0.785}  & \textbf{0.798}  &  0.774 & \textbf{0.763}  &  \textbf{0.763}    \\
			& -	w/o Geographic Knowledge & 0.738   & 0.764  & 0.774 & 0.765  & 0.751    & 0.751   \\
			& - w/o Dependency Relations & 0.731   & 0.778  & 0.772 & \textbf{0.782}  & 0.752    & 0.752   \\
			\bottomrule
		\end{tabular}
	}
	\label{table:main-results}
\end{table*}

\paragraph{Joint Setting}
Although the separate setting is effective, its independent classification decision does not consider the dependencies across output labels. This may result in limiting performance over the task that has strong label dependencies. Thus, we propose a joint setting for this task.

For the joint setting, as illustrated in Figure~\ref{fig:label-schema} and defined in Section~\ref{sec:task_formulation}, we collapse the labels in PTE and PAI into one set for tagging. 
Similar to the separate setting, we use a CRF model to jointly make the tagging decisions. During training, the log-likelihood is as follows:
\begin{align}
\mathcal{L}_\theta = \sum_i\log(p(y|D; \mathbf{\theta}_T)) \,,
\end{align}
where $p(y|D; \mathbf{\theta}_T)$ is the probability function of sequence $y$ in CRF, and $W_T$ is the weight in CRF.
During decoding, we predict the output sequence using the Viterbi algorithm~\cite{forney1973viterbi}.

\section{Experiments}
In this section, we describe experiment settings, evaluating metrics, and report empirical results on the WebPOIs dataset.

\subsection{Settings and Evaluation Metrics}
We use Glove.840B.300d~\cite{pennington2014glove} embeddings as the pre-trained word embeddings. The pre-trained word embeddings are fixed during training. We use Adam~\cite{Kingma2014AdamAM} to optimize our model with the learning rate of 0.001, and the batch size is set to 32. We set the maximum epoch number for training to 10.

For evaluation, we employ the $F_1$ score of each accessibility label and use Macro/Micro $F_1$ as the overall evaluation metrics.
In the separate setting, an accessibility label is regarded as correct when its accessibility type and the corresponding POI mention are both correct.

\subsection{Comparison Methods} 
\label{baselines}
We evaluate GEDIT against the following mainstream methods used in general sequence tagging tasks. 

\begin{itemize}
	\item \textbf{CNN+CRF}~\cite{kim2014convolutional} is a CNN structure on the character or word sequence to learn representations of n-grams from a document for the named entity recognition (NER) task.
	
	\item \textbf{BiLSTM+CRF}~\cite{huang2015bidirectional} is a bidirectional LSTM model to learn representations of words from a document with a CRF model for the NER task.
	
	\item \textbf{LR-CNN}~\cite{gui2019cnn} is a CNN-based method that incorporates lexicons using a rethinking mechanism.
	
	\item \textbf{CNN+BiLSTM+CRF}~\cite{wu2019neural} is a CNN-LSTM-CRF neural architecture to capture both local and long-distance contexts for named entity recognition. The model jointly trains NER and word segmentation models to enhance the ability of the NER model in identifying entity boundaries. 
	
	\item \textbf{ERNIE+CRF}~\cite{Sun2019ERNIEER} uses pre-trained ERNIE model to learn the word representations from a document and uses CRF for POI and accessibility decoding.
\end{itemize}

\subsection{Results and Analysis}
Table~\ref{table:main-results} shows the performance of different models. Results show that GEDIT significantly outperforms all baselines. Specifically, we have the following observations.

(1) The models with joint setting perform better than those with separate setting. It suggests that jointly training PTE and PAI can better utilize the shared information between the two tasks. It also shows the advantages of considering the interactions between the two relevant tasks of PTE and PAI.
    
(2) LSTM-based models (BiLSTM+CRF and CNN+BiLSTM+CRF) perform better than CNN-based models (CNN+CRF and LR-CNN), which indicates that this task is susceptible to the sequence order. CNN is good at extracting local n-gram features from the document. However, it cannot model the word order information well. Furthermore, we can observe that CNN+BiLSTM+CRF performs better than both CNN-based and LSTM-based models, which shows that combining the advantages of CNN and LSTM is able to better model this task.
    
(3) Compared with both CNN-based and LSTM-based models, the models using ERNIE as an encoding module significantly outperform them in both separate and joint settings. It suggests that pre-trained language models have a stronger ability in modeling the semantic representations of sentences than CNN-based and LSTM-based models.
    
(4) GEDIT significantly outperforms all baselines. Compared with the ERNIE-based model, GEDIT further introduces geographic-enhanced text representations. Moreover, it considers the dependency relations of different text nodes, and applies a relational graph convolutional network to encode this kind of relations to overcome the influence of POIs that do not have accessibility changes. As a result, it is able to make more accurate accessibility predictions.
    
(5) Compared with GEDIT with soft attention (GEDIT$_{soft}$), GEDIT with hard attention (GEDIT$_{hard}$) performs better on both separate and joint settings. Hard attention directly uses the node representation that belongs to the sub-word. By contrast, soft attention tries to fuse all graph nodes by the attention, which may fuse more noise and make predictions less accurate than hard attention. However, we can observe that GEDIT$_{soft}$ still works better than ERNIE+CRF, which further shows that encoding the dependency relations of different nodes into sub-word representation helps predict the POI's accessibility.

Overall, our model (GEDIT$_{hard}$) achieves the best performance of 76.3\% in terms of both Macro $F_1$ and Micro $F_1$. 

\subsubsection{Ablation Studies}
We also performed extensive ablation experiments over the two components of GEDIT$_{hard}$ to figure out their relative importance. Specifically, three variations of GEDIT$_{hard}$ with the following settings are implemented for comparison.
\begin{itemize}
	\item \textbf{w/o Geographic Knowledge}: Replace the GERNIE component with the vanilla ERNIE model. 
	\item \textbf{w/o Dependency Relations}: Remove the dependency relation learning component and only use GERNIE.
	\item \textbf{ERNIE+CRF}: Remove both GERNIE and dependency relation learning components, and only utilize the ERNIE model to tag POIs and their accessibility labels.
\end{itemize}

The results of ablation studies are presented at the bottom of Table~\ref{table:main-results}. From the results, we observe that:

(1) Compared with GEDIT$_{hard}$, both the Macro $F_1$ and Micro $F_1$ of ``GEDIT$_{hard}$ - w/o Geographic Knowledge'' model decline by 1.2\% absolutely. This shows that replacing the vanilla ERNIE model with the geographic knowledge enhanced model GERNIE can bring significant improvements to this task. The main reason is that GERNIE is able to relieve the problem of newly-coined and OOV words, which facilitates accurately extracting POI names.

(2) Compared with GEDIT$_{hard}$, both the Macro $F_1$ and Micro $F_1$ of ``GEDIT$_{hard}$ - w/o Dependency Relations'' model decline by 1.1\% absolutely. This shows that introducing dependency relations can also bring significant improvements to this task. The main reason is that dependency relations can help avoid the distraction from the auxiliary POIs that do not have any accessibility changes.

(3) GEDIT$_{hard}$ significantly outperforms ERNIE+CRF by a large margin in terms of all metrics. In addition, GEDIT$_{hard}$ also significantly outperforms both ``GEDIT$_{hard}$ - w/o Geographic Knowledge'' and ``GEDIT$_{hard}$ - w/o Dependency Relations'' in terms of Macro $F_1$ and Micro $F_1$. This demonstrates that incorporating both components of geographic knowledge and dependency relations into a sequence tagging framework for \textit{joint POI and accessibility extraction} can lead to more significant improvements than using neither of them or using either of them individually.

\subsubsection{Performance of POI Term Extraction}
To investigate the effectiveness of different models on POI term extraction, we conduct the experiments and demonstrate the performance in Table~\ref{table:pte-results}. All models in experiments are conducted with the joint setting. 

\begin{table}[!ht]
	\centering
	\caption{Results of POI term extraction.}
	\begin{tabular}{l|rrr}
		\toprule
		\textbf{Model}  & \textbf{Precision} & \textbf{Recall} & \textbf{$F_1$} \\
		\midrule
		\midrule
		\textbf{CNN+CRF}     & 0.648 &  0.600 & 0.623  \\
		\textbf{LR-CNN}      & 0.692 &  0.632 & 0.661  \\
		\textbf{BiLSTM+CRF}  & 0.703 &  0.641 & 0.670  \\
		\textbf{CNN+BiLSTM+CRF}  & 0.711 &  0.681 & 0.696  \\
        \textbf{ERNIE+CRF}     & 0.711 & 0.729 & 0.720 \\
		\textbf{GEDIT$_{soft}$}  & 0.738 & 0.788 & 0.762 \\
		\textbf{GEDIT$_{hard}$}  & \textbf{0.746} & \textbf{0.793} & \textbf{0.769} \\
		\bottomrule
	\end{tabular}
	\label{table:pte-results}
\end{table}

From the results in Table~\ref{table:pte-results}, we observe that:

(1) ERNIE-based models have an obvious advantage over CNN-based and LSTM-based models. The main reason is that the POI terms are usually rare words, making the POI tagging more difficult than the vanilla named entity recognition task.

(2) The proposed model (GEDIT) shows better performance than ERNIE+CRF. The main reason is that combining both sub-word and tree node representations can guide the model to better find the boundaries of POIs.

To obtain a deeper understanding of the effect of POI name length on the performance of our model, we further compare the performance of different models on POIs with various lengths. To accomplish this, we separate the POI set into three groups: (1) POIs with 3 or less ($\leq$3) Chinese characters (short); (2) POIs with 4--5 ($>$3 \& $\leq$5) Chinese characters (medium); and (3) POIs with $>$5 Chinese characters (long). We report the performance of GEDIT$_{hard}$ and ERNIE on extracting short, medium, and long POIs.

\begin{table}[h]
\centering
\caption{Performance on different groups of POIs.}
\begin{tabular}{@{}l|crrr@{}}
\toprule
\textbf{Model}                                           & \textbf{Group} & \textbf{Precision} & \textbf{Recall} &  \textbf{$F_1$} \\ \midrule
\midrule
\multirow{3}{*}{\textbf{ERNIE+CRF}} 
& Short  & 0.638 & 0.637 & 0.638  \\
& Medium & 0.722 & 0.747 & 0.734  \\
& Long & 0.719 & 0.741 & 0.730       \\
\midrule
\multirow{3}{*}{\textbf{GEDIT$_{hard}$}}
& Short & 0.687 & 0.722 & 0.704 \\
& Medium & 0.747 & 0.817 & 0.780 \\
& Long & 0.731 & 0.799 & 0.764 \\ \bottomrule
\end{tabular}
\label{table:poi_term_length}
\end{table}

Table~\ref{table:poi_term_length} shows the results. We observe that GEDIT$_{hard}$ achieves 6.6\% (short), 4.6\% (medium), and 3.4\% (long) absolute improvements over ERNIE+CRF in terms of $F_1$, which demonstrates that GEDIT$_{hard}$ achieves greater improvements for shorter POIs. The main reason is that shorter POIs convey less information than longer ones, making it more difficult for a model to learn from. Therefore, the shorter a POI is, the more difficult it is to make an accurate prediction. GEDIT$_{hard}$ is able to utilize external geographic knowledge to mitigate this issue, and consequently performs better on shorter POIs.

\section{Practical Applicability}
We describe how we deploy GEDIT into the POI data maintenance process at Baidu Maps. In each week, we first extract millions of documents containing POIs from multiple data sources, including general Web documents and official websites. Next, we filter these documents by a list of keywords that indicate POI accessibility changes, which generates hundreds of thousands of documents. Then, we feed these documents into GEDIT. Once we obtain the extracted <POI name, accessibility label> pairs from GEDIT, we use some heuristic rules to remove the inappropriate pairs. The pairs are then sent to a linking process to decide whether we add the extracted POI into the POI database. For those pairs that have an accessibility label of NEW, the linked ones are abandoned. For those pairs that have other accessibility labels, we abandon the un-linked ones. 
After finishing all the procedures described above, we can obtain about 4,000 to 10,000 <POI name, accessibility label> pairs per week. These pairs are finally sent to the operators for manual verification to ensure that the data are accurate and compliant at Baidu Maps.

A key indicator of the effectiveness of the POI data maintenance is the success rate of manual verification (SRMV). SRMV means that of all the <POI name, accessibility label> pairs sent for verification, how many of them are eventually confirmed true and accepted for publication in the POI database.
The quality of extracted <POI name, accessibility label> pairs will directly affect SRMV.
After we deployed the GEDIT model into the POI data maintenance process, the SRMV increases by 17.8\% compared to the previously deployed extractor.
This demonstrates that GEDIT is able to save significant human effort and labor costs, which confirms that it is a practical solution for POI accessibility maintenance.

\section{Discussion}
We first discuss an alternative way to accomplish the task.
A natural idea to accomplish this is to directly train a binary classifier to check whether a document contains POI accessibility information, and then send the document with accessibility labels to an operator for manual verification.
Although this is a straightforward method, it usually takes an operator a lot of time on identifying the boundaries of a POI, and thus can hardly be applied in large-scale POI data maintenance.
This is evident from the building of the WebPOIs dataset. During which, we found that the most time-consuming stage of manual annotation comes from the annotation of POI boundaries. It becomes even worse when the operators are not familiar with the POI names. As a consequence, they inevitably take a lot of time on determining the boundaries of POIs by repeatedly verifying and searching for additional information on the Internet, which would greatly reduce the work efficiency. 

Moreover, the accessibility of a POI strongly correlates with the time-dependent variation of individual business activities,
which necessitates extracting <POI name, accessibility label, time of taking effect> triplets rather than only producing <POI name, accessibility label> pairs from unstructured text.
In practice, the deployment of GEDIT is accompanied by a post-processing step, which applies a heuristic method to extract time information.
Specifically, the accessibility time of a POI is obtained by: (1) extracting the time entity in the document with a NER model and (2) using the document creation time if the previous method fails.
However, it remains challenging to identify the accurate accessibility time (e.g., ``March 31, 2020'') of a POI due to loosely described date and time (e.g., ``Mother's Day'' and ``September 9th in lunar calendar''), fuzzy date and time (e.g., ``opening soon'' and ``around the Mid-Autumn Festival in 2021''), and relative time span (e.g., ``the first day of the Dragon Boat Festival holiday'' and ``three days later'').
In addition,  there may exist multiple <POI name, accessibility label, time of taking effect> triplets in one document, which brings a major challenge to map each time chunk to the corresponding POI chunk.
To address these challenges, we are developing an end-to-end, time-aware extension of GEDIT, which will be introduced in our following work.

\section{Related Work}
Here we briefly review the closely related work in the fields of POI maintenance and named entity recognition.

\subsection{POI Maintenance}
POI database maintenance, which includes discovering emerging POIs and updating existing POIs, is an essential and crucial task for commercial map applications. In the past, this procedure has generally relied on manual input, which is tedious and expensive~\citep{Mummidi2008DiscoveringPO,Ruta2012SemanticAO}. With the emerging of massive user-generated content and the development of machine learning methods, several methods have been proposed to effectively discover or update POI information.

Some early studies focus on discovering POIs from images by recognizing the logs or brand symbols that make the POIs identifiable. Early work proposed to detect logos from images with hand-crafted features~\citep{Revaud2012CorrelationbasedBF,Romberg2011ScalableLR}. With the development of deep learning, neural network-based methods \citep{Su2017WebLogo2MSL,Su2016DeepLL} outperform the previous hand-crafted approaches.
For updating existing POIs, \citet{Revaud2019DidIC} proposed to detect changes of POIs based on comparing two image sets of the same venue at different times.

Several work has attempted to leveraging text data such as Web snippets~\citep{Rae2012MiningTW,Chuang2016EnablingMS}, yellow pages~\citep{Ahlers2013BusinessER}, and tweets~\citep{Xu2019DLocRLAD} to discover emerging POIs.
For updating existing POIs,
\citet{Zhou2013APD} proposed a method for updating POIs based on Sina Weibo check‐in data.
They detect new POIs by analyzing the check-in data that emerges over time.
\citet{Chuang2018DetectingOP} proposed a feature-based method for detecting outdated POIs using crawled Web snippets. 

Our work is significantly different in the following aspects. 
(1) We consider a complete task of POI database maintenance, including discovering new POIs and updating existing POIs, while other studies only focus on a portion of the task.
(2) We take advantage of both pre-trained language models and dependency parsing to guide a sequence tagging model, which is able to jointly extract POI mentions and identify their coupled accessibility labels from unstructured text, saving much labor costs in practice.

\subsection{Named Entity Recognition}
Since POI terms can be regarded as a kind of named entity, one of our sub-tasks, POI term extraction, is closely related to named entity recognition (NER), which is a traditional natural language processing task. 
NER is regarded as a sequence tagging problem. Early studies used feature-based classifiers~\citep{Lafferty2001ConditionalRF,Florian2003NamedER} to build a tagger model.
With the development of neural networks, studies based on CNN~\citep{Collobert2011NaturalLP} and LSTM-CRF~\citep{Chiu2015NamedER,Lample2016NeuralAF,Ma2016EndtoendSL} get promising results. Recently, fine-tuned models based on pre-trained language models such as ELMO~\citep{Peters2018DeepCW} and BERT~\citep{Devlin2019BERTPO}, have achieved impressive performance.

The difference between our proposed task and NER lies in that the main evidence used to predict the output tag is different.
In NER the output tag of an entity's type such as ORG or PER is mainly determined by its name, while in our study the output tag of accessibility is determined by the context.

\section{Conclusions and Future Work}
It is of vital importance to provide timely accessibility reminders of POIs to the users at commercial map applications.
In this paper, we present a novel task that jointly extracts POI mentions and identifies their coupled accessibility labels from unstructured text.
We formulate it as a sequence tagging problem, where the goal is to produce <POI name, accessibility label> pairs from unstructured text. To address the two challenges: (1) rare or unknown words and (2) many-to-one or one-to-many <POI name, accessibility label> mapping, we propose a \textbf{G}eographic-\textbf{E}nhanced and \textbf{D}ependency-gu\textbf{I}ded sequence \textbf{T}agging (\textbf{GEDIT}) model. GEDIT not only adopts a geographic-enhanced pre-trained model to learn the text representations, but also applies a relational graph convolutional network to learn the tree node representations from the parsed dependency tree.
Extensive experiments conducted on a real-world dataset demonstrate the superiority and effectiveness of GEDIT.
In addition, statistics show that the proposed solution can save significant human effort and labor costs to deal with the same amount of documents.

In the future, we consider extending the proposed solution and further addressing the following open problems.

(1) Other attributes, such as the exact time that indicates when a POI changes its accessibility label,  also deserve to be extracted from Web text. We plan to explore ways to identify and extract such attributes in the future.

(2) The users' search \citep{huang2016generating,huang2017learning,huang2020multi,p3ac-kdd20,p4ac-kdd21,10.1145/3447548.3467059} and navigation \citep{fang2020kddeta,fang2021kddeta} behaviors on visiting opened POIs differ from those on visiting closed POIs, which can be leveraged as valuable evidence to detect changes of POIs.
As future work, we plan to investigate whether it is practical to identify accessibility changes of POIs from the search and navigation logs of map applications.

(3) The accessibility of POIs obtained from Web text needs to be further verified by human annotations. We plan to develop solutions to automatically perform verification and validation steps, which significantly reduces labor costs.


\balance
\bibliographystyle{ACM-Reference-Format}
\bibliography{main}

\end{document}